\documentclass[wcp]{jmlr}

\usepackage{longtable}

\usepackage{booktabs}
\usepackage{times}
\usepackage{latexsym}
\usepackage{graphicx}
\usepackage{multirow}
\usepackage{amsfonts,amsmath,amssymb,bm}
\usepackage{graphicx}
\usepackage{xcolor}
\usepackage{setspace}
\usepackage{pdflscape}
\usepackage{enumitem}
\usepackage{rotating}
\usepackage{caption}
\usepackage{subcaption}
\usepackage{wrapfig}

\usepackage{lineno}

\makeatletter
\let\Ginclude@graphics\@org@Ginclude@graphics 
\makeatother

\jmlrvolume{}
\jmlryear{2024}
\jmlrworkshop{ACML 2024}

\title[QOG:Question and Options Generation based on Language Model]{QOG:Question and Options Generation\\ based on Language Model}

 \author{\Name{Jincheng Zhou} \Email{zhoujincheng@gmail.com}\\
 \addr Zhuoshi Technology,  University of Electronic Science and Technology of China
 \AND
 \Name{Yue Hu} \\
 \addr Zhuoshi Technology
 \AND
\Name{Ya Wang} \\
\addr Zhuoshi Technology
}

\editors{Vu Nguyen and Hsuan-Tien Lin}

\begin{document}

\maketitle

\begin{abstract}
Question-Options Generation (QOG) is a task that involves generating a set of question-options pairs within a given input. 
This task has various applications, including fine-tuning large models, information retrieval, and automated multiple-choice question generation for education.
In this paper, we develop QOG models using three different methods based on fine-tuning sequence-to-sequence language models (LMs). 
Experiments demonstrate that the end-to-end QOG model is computationally efficient and stable during both training and inference, 
outperforming other methods.
Furthermore, our analysis indicates that our QOG models are competitive on the QOG task compared to the large language model Llama 3-8B.
\end{abstract}

\section{Introduction}
Question-Options (QO) is derived from Question-Answering (QA), extending question-answer pairs by adding three text-related distractors. 
Compared to QA, QO provides more information by combining the correct answer with several specially designed incorrect answers. 
This information enables the model to incorporate more knowledge and improve its ability to identify potential errors. 
When used for fine-tuning large language model\citep{hu2021lora,liu2021p,liu2023gpt}, the goal is to identify the correct answer from four candidate answers, 
which improves the model's understanding of the question and its ability to distinguish the differences between different options, 
making the model understand the text better.
When used for model evaluation, 
whether the model can select the correct answer becomes a key indicator of its text learning and processing capabilities
\citep{hendrycks2020measuring,huang2024c,cobbe2021training}.

Question and Options Generation (QOG) refers to the task of generating a set of question-options pairs given an input context 
(e.g. a paragraph). 
QOG can be used to develop unsupervised question answering and as a data augmentation tool
\citep{alberti2019synthetic,yu2018fast,riabi2020synthetic,singh2019xlda,longpre2019exploration} 
to enhance the text understanding capabilities of large language models. 
QOG can also be used as an educational aid\citep{agarwal2019eduqa,cai2023paniniqa,wang2023mcqa},
to enhance information retrieval models\citep{nogueira2019document,pyatkin2021asking}, 
and as a way to explain models.

\begin{figure}[t]
    \centering
    \includegraphics[width=1\columnwidth]{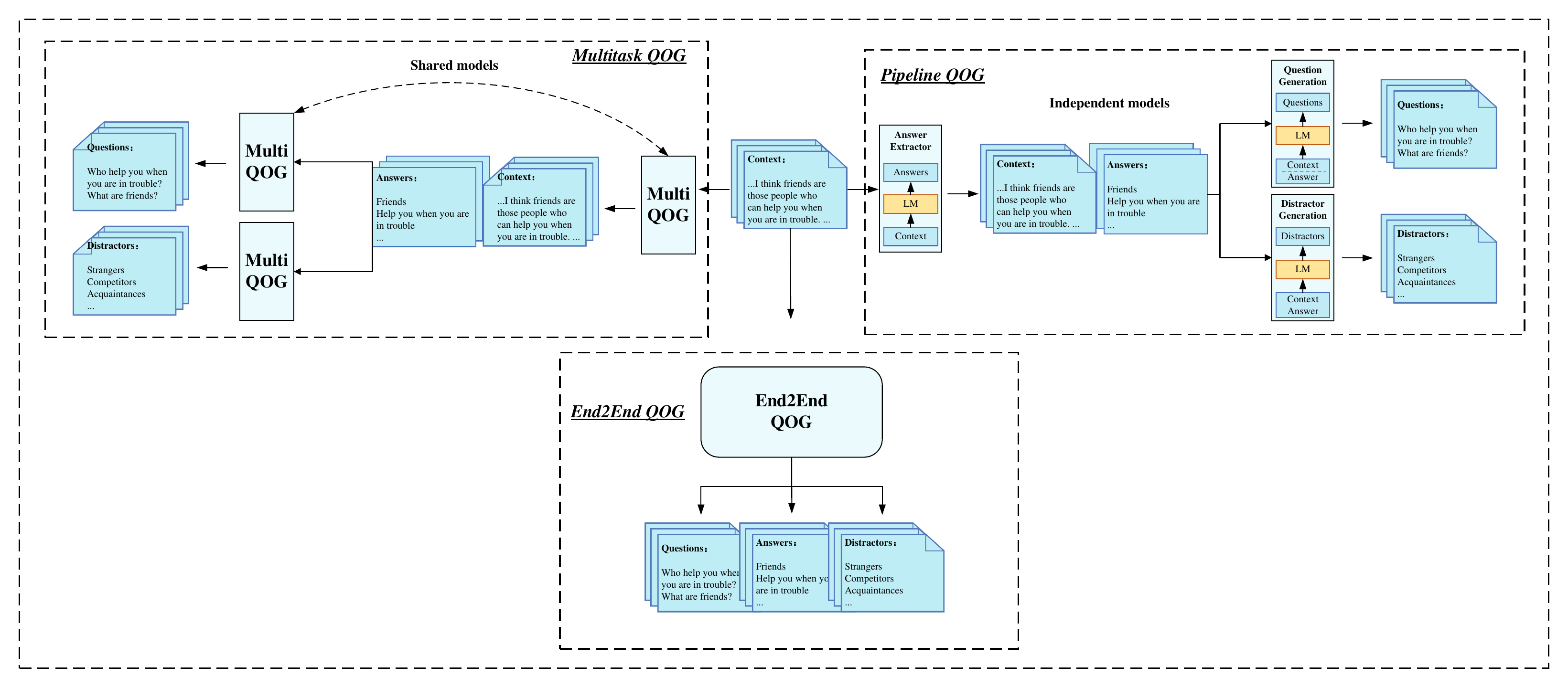}
    \caption{Overview of the considered QOG approaches.}
    \label{fig:overview}
\end{figure}

QOG originates from Question Generation (QG)\citep{duan2017question,du2017learning,ali2010automatic}, 
which involves generating a question for an answer given an input context. 
Compared to QG and QAG, QOG presents a more intricate challenge, as the correct answer and distractors must be constructed rather than assumed as part of the input.
Currently, it is unknown which models are effective for the QOG task because no comprehensive comparative study has been established to date.

In this paper, we consider the QOG as a task of generating questions and options given a context, 
and compare three simple QOG methods based on fine-tuning encoder-decoder language models such as T5\citep{T5}.
The three methods are common methods for fine-tuning language models now: (1) Pipeline QOG, which decomposes the task into three parts: 
answer extraction, question generation and distractor generation, and learns a separate model for each subtask;
(2) Multi-task QOG, which uses a shared single model to train the three subtasks simultaneously instead of separate models; 
and (3) End-to-end QOG, which directly generates questions-options using end-to-end sequence-to-sequence learning.
Finally, we introduce GPT-4 as a referee based on traditional evaluation methods and compare the three methods in multiple domains to 
objectively evaluate the cross-domain generalization ability of the models.

\section{Related Work}
There is no work using pre-trained LM for QOG, but there are related works on QG and QAG.
\subsection{Question Generation}
Question generation (QG) aims to generate questions given documents and answers. 
This approach has been applied in a variety of scenarios, including data augmentation\citep{alberti2019synthetic,pellicer2023data}
and document retrieval\citep{nogueira2019document,hofstatter2020local,ram2023context}.
\citet{puri2020training} fine-tuned QG's autoregressive LM to generate related questions. 
These works focus on generating large-scale data at once without paying much attention to data quality.
\citet{xiong2020ta} followed the idea of adversarial generation and used question-answering agents and QG models to train adversarial models. 
This work used semantic similarity as an evaluation indicator to improve the quality of generated data, 
but did not focus on the generalization ability of the model in other fields.
To this end, our work will focus on improving data quality while testing the generalization ability of the model with data sets from different fields.

\subsection{Question-Answer Generation}
Question-Answer Generation (QAG) generates question-answer pairs based on a given document. 
QAG can be used to develop QA models without human supervision and as an auxiliary tool for data augmentation of large language models.
\citet{bartolo2021improving} used QAG model to generate adversarial examples for QA,
\citet{lewis2021paq} improved extractive QA by generating millions of question-answer pairs through QAG,
In both examples, the fine-tuned model is BART.
In order to improve the recall rate of QA, \citet{fang2020accelerating} regarded answer extraction as a classification problem and used RoBERTa\citep{liu2019roberta} to implement it.
\citet{qasnowball} generated QA iteratively and improved the data quality of each iteration by modifying the seed set.
\citet{ushio2023empirical} made a comparison of several QAG methods and established a complete system.
These works can generate high-quality QA pairs, however they are limited to machine metrics and do not evaluate QA quality according to human standards.

\subsection{QG \& QAG Evaluation}
In terms of evaluating the quality of QG, \citet{zhang2019addressing} believed that the traditional evaluation criteria were not accurate enough, 
they proposed using the QG model to generate questions for each answer in the data set, and then using the synthetic QA pairs to train the QA model. 
The performance of these QA models will indirectly measure the ability of the QG model.
\citet{sultan2020importance} also emphasized the limitations of traditional QG evaluation methods and 
introduced new indicators for evaluating diversity to better evaluate the quality of QG.
These studies enrich the evaluation system by introducing new evaluation criteria. However, these criteria still have certain deviations from human evaluation criteria.
Shortly after the release of GPT-4, \citet{moore2023assessing} proposed a method for evaluating the quality of multiple-choice questions based on GPT-4,
with a correlation of more than 80\% with human evaluation.
In this paper, we also choose to introduce GPT-4 as part of the evaluation system to make the generated QO more in line with human evaluation standards.

\section{Question \& Options Generation} \label{sec:methodology}
Given a context $c$ (e.g. a paragraph), the goal of QOG is to generate question-options pairs that are relevant to the information in $c$:
$\mathcal{QO}_c = \{ (q^1, o^1), (q^2, o^2), \dots \}$.
In the following, we will introduce three different QOG methods based on fine-tuned language models.

\subsection{Pipeline QOG}\label{pipeline} 
This approach decomposes the QOG task into three simpler subtasks: answer extraction (AE), question generation (QG) and distractor generation (DG).
The AE, QG and DG models are designed to learn from each sample containing context, sentences, answers
and corresponding questions, thereby optimizing their abilities to generate relevant questions, distractors, and extract correct answers.

The AE model $P_{{\rm ae}}$ first generates an answer $a$ in a given context $c$, 
and then the QG model $P_{{\rm qg}}$ generates a question $q$ based on the context $c$ and the answer $a$, 
which can be answered by the answer $a$. Finally, the DG model $P_{{\rm dg}}$ generates distractors $d$ based on the context $c$ and the answer $a$.
The AE, QG, and DG models can be trained independently on any dataset consisting of triples $(c, a, q)$ by maximizing the conditional log-likelihood iteration:
\begin{align}
        \tilde{a} &= argmax_{a} P_{{\rm ae}}(a|c, s) \label{eq:answer-extraction} \\
        \tilde{q} &= argmax_{q} P_{{\rm qg}}(q|c, s, a) \label{eq:question-generation}\\
        \tilde{d} &= argmax_{d} P_{{\rm dg}}(d|c, s, a) \label{eq:distractor-generation}
\end{align}

Similar to other sequence-to-sequence learning, the log-likelihood here is based on token-level predictions. When learning, the input to the AE model is in the form of:
\begin{equation*}
[ c_1, c_2, \dots, c_{|c|} ]
\end{equation*}
where $c_i$ is the $i$-th token in context $c$, and $|n|$ represents the number of tokens in the text.
When input to the QG model, the answer $a$ is considered and the position of the answer $a$ in the context is marked with \texttt{<hl>}. 
The form is:
\begin{equation*}
[ c_1, \dots, \texttt{<hl>}, a_1, \dots, a_{|a|}, \texttt{<hl>}, \dots, c_{|c|}  ]
\end{equation*}
where $a_i$ is the $i$-th token in the answer $a$. The model will learn the pattern of generating a question from the highlighted answer. 
Finally, the answer and context are input into the model $P_{{\rm dg}}$:
\begin{equation*}
  [ c_1, \dots, \texttt{<hl>}, a_1, \dots, a_{|a|}, \texttt{<hl>}, \dots, c_{|c|}  ]
\end{equation*}
At inference time, we replace the input answer $a$ to the QG(\ref{eq:question-generation}) and DG(\ref{eq:distractor-generation}) models during 
training with the prediction result $\tilde{a}$ of the AE model(\ref{eq:answer-extraction}),
and then predict the context $c$ to obtain the generated question-options. Since the result $\tilde{a}$ is used as the input of $P_{{\rm qg}}$ and $P_{{\rm dg}}$,
the effect of this method almost depends on $P_{{\rm ae}}$.

\subsection{Multitask QOG}\label{multitask}
The method mentioned in \ref{pipeline} trains an independent model for each subtask. 
Instead of this, Multitask QOG adopts a multi-task learning approach to fine-tune a shared model for AE, QG, and DG at the same time. 
To be precise, we mix the training data of AE, QG and DG together, and a random batch is extracted at each fine-tuning iteration.
Each subtask is distinguished by adding a task prefix at the beginning of the input text:
``$\texttt{extract answer}$'' (AE), ``$\texttt{question generation}$'' (QG), and ``$\texttt{distractor generation}$'' (DG).
The loss function will combine the loss functions of AE, QG, and DG, and the total loss is the weighted sum of the three:
\begin{equation}
  L=\alpha L_{AE}+\beta L_{QG}+\gamma L_{DG}
\end{equation}
where $\alpha$, $\beta$, and $\gamma$ are hyperparameters. 
This design enables the model to jointly learn three different but related task modes at once, 
thereby improving the overall generalization ability and efficiency.

\subsection{End2end QOG}\label{end2end} 
This method directly inputs the context $c$ and outputs the text containing all question-options pairs.
We can model this by converting the question-options pairs in the dataset into a flattened sentence $y$,
and fine-tuning a sequence-to-sequence model to generate $y$ based on the context $c$.
The format of $y$ is as follows:
\begin{align}
    \mathcal{T}(\mathcal{Q}_c) = ``\texttt{question:}q_1,\texttt{options:}o_1\texttt{|} 
    \texttt{question:}q_2,\texttt{options:}o_2\texttt{|}\dots\textrm'\textrm' \label{eq:prompting}
\end{align}
where each pair of question-options is converted to the format of $y$ and connected by the separator '\texttt{|}'.
The end-to-end QOG model $P_{{\rm qog}}$ is optimized by maximizing the following conditional log-likelihood:
\begin{align}
    \tilde{y} = argmax_{y} P_{{\rm qog}}(y|c) \label{eq:qog}
\end{align}

\section{Evaluation}\label{sec:experiments}

\subsection{Experimental Setting} \label{sec:experimental-setting}

\noindent \textbf{Data.} 
The data for training QOG is a multiple-choice question dataset generated based on SQuAD. 
We send each pair of QA in SQuAD to a large language model (such as GPT-4) to generate distractors to form the dataset SQuAD-QO.
For SQuADShifts and FinQA, we generated the test sets SQuADShifts-QO and FinQA-QO in the same way.
The relevant datasets have been publicly released on \href{https://huggingface.co/datasets/Csking/QO-squad}{HuggingFace}.

\begin{table}[h]
    \centering
    \renewcommand{\arraystretch}{1.5}
    \resizebox{0.8\textwidth}{!}{
    \begin{tabular}{@{}l@{\hspace{3pt}}p{3cm}@{\hspace{5pt}}p{2cm}@{\hspace{5pt}}p{2cm}@{\hspace{5pt}}p{2cm}@{\hspace{5pt}}p{2cm}@{\hspace{5pt}}p{2cm}@{\hspace{5pt}}p{2cm}@{}}
    \toprule
    \multicolumn{2}{@{}l}{Approach}  & Aver         & Amazon     & Wiki        & NYT         & Reddit     & Fin \\\midrule
    \multicolumn{2}{@{}l}{\textit{Llama 3-8B}}                & \textit{49.79} & \textit{51.34} & \textit{47.26} & \underline{\textbf{\textit{59.02}}} & \textit{49.92} & \textit{41.43} \\\midrule
    \multirow{4}{*}{\rotatebox{90}{T5\textsubscript{SMALL}}}
                        & Pipeline  & 42.99 & 42.74 & 40.47 & 42.76 & 41.12 & \textbf{51.90} \\
                        & Multitask & 46.03 & \textbf{50.40} & 41.92 & 43.68 & 46.11  & 48.03\\
                        & End2end   & \textbf{47.98} & 48.41 & \textbf{44.72} & \textbf{48.47} & \textbf{47.48}  & 50.84\\\midrule
    \multirow{4}{*}{\rotatebox{90}{T5\textsubscript{BASE}}}
                        & Pipeline  & 46.45 & 48.52 & 43.46 & 44.43 & 45.53  & \textbf{55.29}\\
                        & Multitask & 47.05 & 47.84 & 45.84 & 51.08 & 44.56  & 45.93\\
                        & End2end   & \textbf{50.78} & \textbf{49.90} & \textbf{47.94} & \textbf{52.14} & \textbf{51.14}  & 52.77\\\midrule
    \multirow{4}{*}{\rotatebox{90}{T5\textsubscript{LARGE}}}
                        & Pipeline  & 51.45 & 50.73 & 48.42 & 46.27 & 49.82 & \underline{\textbf{62.01}} \\
                        & Multitask & 51.61 & 48.95 & 49.22 & 49.19 & 52.65 & 58.04 \\
                        & End2end   & \underline{\textbf{54.66}} & \underline{\textbf{54.58}} & \underline{\textbf{51.28}} & \textbf{55.56} & \underline{\textbf{52.16}} & 59.71\\\bottomrule
    \end{tabular}
  }
    \caption{QO evaluation results($F_1$) of different QOG models on the test set. For comparison, we introduced Llama 3-8B.
    The best score of QOG methods in each LM is shown in bold, and the best result in each domain across all models is underlined.}
    \label{tab:mqae}
\end{table}

\noindent \textbf{Evaluation.}
Since QOG's output involves a variety of questions and options, traditional natural language generation metrics are not applicable to it.
In order to evaluate the quality of generated questions and options (QO) comprehensively, we adopted the following two methods:
1) We randomly select 100 questions in the generated QO dataset and call GPT-4 to answer them, the more questions answered correctly indicates the higher quality of the generated QOs, where high quality only indicates answerability, as we noticed that the model sometimes generates questions with unclear meanings or fails to generate them properly.
2) We generate a test set with a different dataset than SQuAD and use the F1 scores to assess the generalisation ability of the QOG model to other domains.
For this purpose, we choose two datasets, SQuADShifts, which covers English reading comprehension tasks in four domains (Amazon/Wiki/Nyt/Reddit), 
and FinQA, which is a Q\&A dataset in the financial domain.
They cover Q\&A in different domains and can effectively judge the applicability and flexibility of the model within each domain.

\begin{wraptable}{r}{0.325\textwidth}
    \centering
    \vspace{-1em}
    \resizebox{0.325\textwidth}{!}{
        \begin{tabular}{ll}
            \toprule
            \textbf{Dataset} & \textbf{Size} \\
            \midrule
            SQuAD-QO & 87,399 \\
            SQuADShifts-QO & 37,691 \\
            FinQA-QO & 8,281 \\
            \bottomrule
        \end{tabular}
    }
    \caption{Size of Datasets}
    \label{tab:dataset_split}
    \vspace{-1.2em}
\end{wraptable}

\noindent \textbf{T5 \& Llama 3.}
For the three approaches mentioned above (i.e., pipelined, multitasking, and end-to-end), we use T5 as the base language model for our experiments.
The model weights used include T5-\textsubscript{small,base,large}, which are all open source on the HuggingFace platform. 
In addition, we report the results of the latest open-source large model Llama 3-8B as a QOG model and compare it with T5.

\subsection{Results} \label{sec:result}

\begin{figure*}[!t]
    \centering
    \includegraphics[width=1\columnwidth]{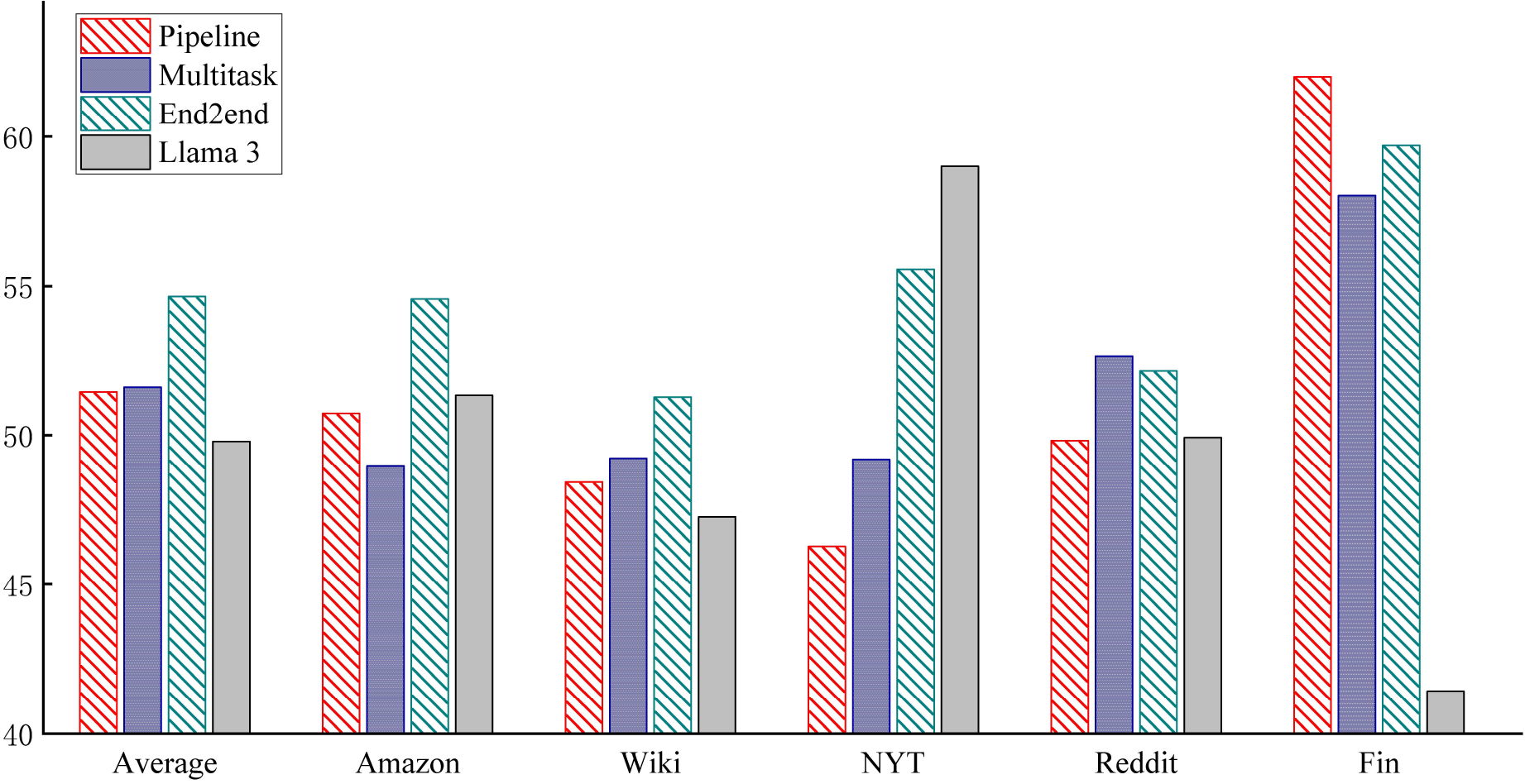}
\caption{QO quality assessment results ($F_1$ scores, 95\% confidence intervals) generated by 
T5\textsubscript{LARGE} multitask/pipeline/end2end on different domain datasets, compared with Llama 3-8B.}
    \label{fig:downsample}
\end{figure*}

\autoref{tab:mqae} shows the evaluation results for the three methods considered. From T5\textsubscript{SMALL} to T5\textsubscript{LARGE}, 
the scores are improving for either method, showing that the model size has a significant positive impact on the performance.
Best model among all models T5\textsubscript{LARGE}(End2end)
achieves the best results in three of the five domains and outperforms Llama 3-8B on average. Even smaller models, such as T5\textsubscript{SMALL},
generate QO pairs of decent quality.

Given the results, the End2end approach achieved the best results on the vast majority of domains, with the highest average scores on T5 models of all sizes. 
Our analysis is that the effect of the Pipeline method depends on the first AE model, which passes its inputs backwards, and the overall effect is actually
is decreasing. The Multitask method shares the same neural network, so the output of the three tasks is more stable.
The problem with it is that only one third of the parameters are available for each task on average, which makes the model's performance degrade.
The advantage of End2end is that it treats the QOG as a single task,
all parameters are updated during training to optimise this task, which gives the model the best generalisation and performance of the three.

\begin{wraptable}{r}{0.325\textwidth}
    \centering
    \vspace{-1em}
    \scalebox{0.9}{
    \begin{tabular}{@{}lc@{}}
    \toprule
    \textbf{Approach}                     & \textbf{GPT-4} \\\midrule
    Llama 3-8B                            & \underline{\textbf{94}}    \\\midrule

    T5\textsubscript{SMALL} (pipeline)    & 64                         \\
    T5\textsubscript{SMALL} (multitask)   & 53                         \\
    T5\textsubscript{SMALL} (end2end)     & 61                         \\
    T5\textsubscript{BASE} (pipeline)     & 79                         \\
    T5\textsubscript{BASE} (multitask)    & 68                         \\
    T5\textsubscript{BASE} (end2end)      & 76                         \\
    T5\textsubscript{LARGE} (pipeline)    & 85                         \\
    T5\textsubscript{LARGE} (multitask)   & 77                         \\
    T5\textsubscript{LARGE} (end2end)     & 83                         \\
    \bottomrule 
    \end{tabular}
}
    \caption{Mean scores for each QOG model under the GPT-4 judgement}
    \vspace{-1.5em}
    \label{tab:GPT-4 evaluation}
\end{wraptable}

We also notice that the Pipeline method performs optimally on FinQA, because the answers of FinQA are mainly numbers of short length (e.g. 236), 
and the AE model focuses on this while learning and obtains the most efficient extraction pattern.
This brings us some inspirations: If the data of QOG has some regularity, we can use the Pipeline method to train the model to get better performance.

\subsection{Assessing the quality of QO pairs using GPT-4}
\label{sec:gpt_evaluation}
GPT-4 is currently the best large language model and the one that best meets human evaluation criteria.
In order to comprehensively evaluate the quality of the QO pairs generated by each QOG model, we use GPT-4 as the judging model.
The specific method is to have each QOG model generate 100 QO pairs in different fields, and then GPT-4 answers these QO pairs.
If GPT-4 answers correctly, it means that the question-options are logical and answerable, which indicates that these QO pairs are qualified.
In this way, we performed a comprehensive evaluation of the quality of the QOs generated by each model, and the results are shown in \autoref{tab:GPT-4 evaluation}.

\begin{wraptable}{r}{0.325\textwidth}
    \centering
    \vspace{-0.5em}
    \renewcommand{\arraystretch}{1.5}
    \scalebox{0.7}{
    \begin{tabular}{@{}l@{\hspace{3pt}}l@{\hspace{5pt}}c@{\hspace{5pt}}c@{\hspace{5pt}}c@{\hspace{5pt}}c@{\hspace{5pt}}c@{\hspace{5pt}}c@{}}
    \toprule
    \multicolumn{2}{@{}l}{\textbf{Approach}}  & \textbf{Compute}(ms)         & \textbf{Memory}(MB)   \\\midrule
    \multirow{4}{*}{\rotatebox{90}{T5\textsubscript{SMALL}}}
                        & Pipeline  & 363.62 & 1176.38  \\
                        & Multitask & 105.94  & 696.95 \\
                        & End2end   & 142.14 & 695.45  \\\midrule
    \multirow{4}{*}{\rotatebox{90}{T5\textsubscript{BASE}}}
                        & Pipeline  & 909.23 & 3053.30 \\
                        & Multitask & 217.75 & 1336.99 \\
                        & End2end   & 301.83 & 1336.28 \\\midrule
    \multirow{4}{*}{\rotatebox{90}{T5\textsubscript{LARGE}}}
                        & Pipeline  & 1480.73 & 8960.01\\
                        & Multitask & 1597.59 & 3310.34 \\
                        & End2end   & 724.96 & 3314.11  \\\bottomrule
    \end{tabular}
}
    \caption{Inference time and memory usage of each QOG model (CPU environment, results are averaged over 100 experiments)}
    \vspace{-2.5em}
    \label{tab:cost}
\end{wraptable}

We found that Llama 3-8B, which previously performed mediocrely in $F_1$ scores, achieved the best results in all areas under the judgement of GPT-4, 
indicating that the QO generated by Llama 3 is superior to other models in terms of logic.
One possible explanation is that Llama 3 does not have a fixed way of generating QO, while T5, which we have fine-tuned, 
has a fixed way of generating QO, resulting in a higher score when calculating similarity.
This result shows that it is not enough to evaluate only from the $F_1$ score,
the method we introduced GPT-4 to evaluate will provide a valuable quality indicator for the QO generated by the QOG model.

Although the proposed model did not outperform the Llama 3-8B, there is one positive outcome: the fine-tuned best model, T5\textsubscript{LARGE}, 
remains competitive under this evaluation and can accurately generate QOs that meet human requirements.

\subsection{QOG Model Comparison}\label{sec:comparison}

In this paper, we have compared the performance of three QOG methods. However, performance is not the only criterion to be considered when choosing a QOG model, 
as each method has its own advantages and limitations in terms of computational cost and usability.
We conducted experiments on each QOG model in the same environment to verify their inference time and memory usage, and the results are shown in \autoref{tab:cost}.

In terms of computational resources consumed, the End2end method outperforms the other two methods because it can generate QO pairs in a single inference process. 
In contrast, both the Pipeline and Multitask methods require a total of three inference sessions to generate QO pairs.
The computational cost of all three methods is proportional to the number of tokens in the input text.
From the perspective of memory usage, both Multitask and End2end methods use only one model to complete the task, 
while the Pipeline method consists of three models, which is three times as much as the other two in terms of memory.

Finally, the performance of the Pipeline method depends on the ability of answer extraction and is limited by the fact that 
the three models are independent of each other in terms of their consistency with respect to the objective, 
although it performs very well on datasets with regular answers.
The Multitask method shares a single model, which allows consistency between the three tasks to be improved, 
however each task gets only one third of the parameters on average, which is a limitation of its performance.
The End2end method treats the QOG as a whole task and therefore it performs better than the other methods in generating QO pairs.

\begin{figure}[t]
    \centering
    \begin{minipage}[b]{0.45\textwidth}
        \centering
        \includegraphics[width=\textwidth,height=0.8\textwidth]{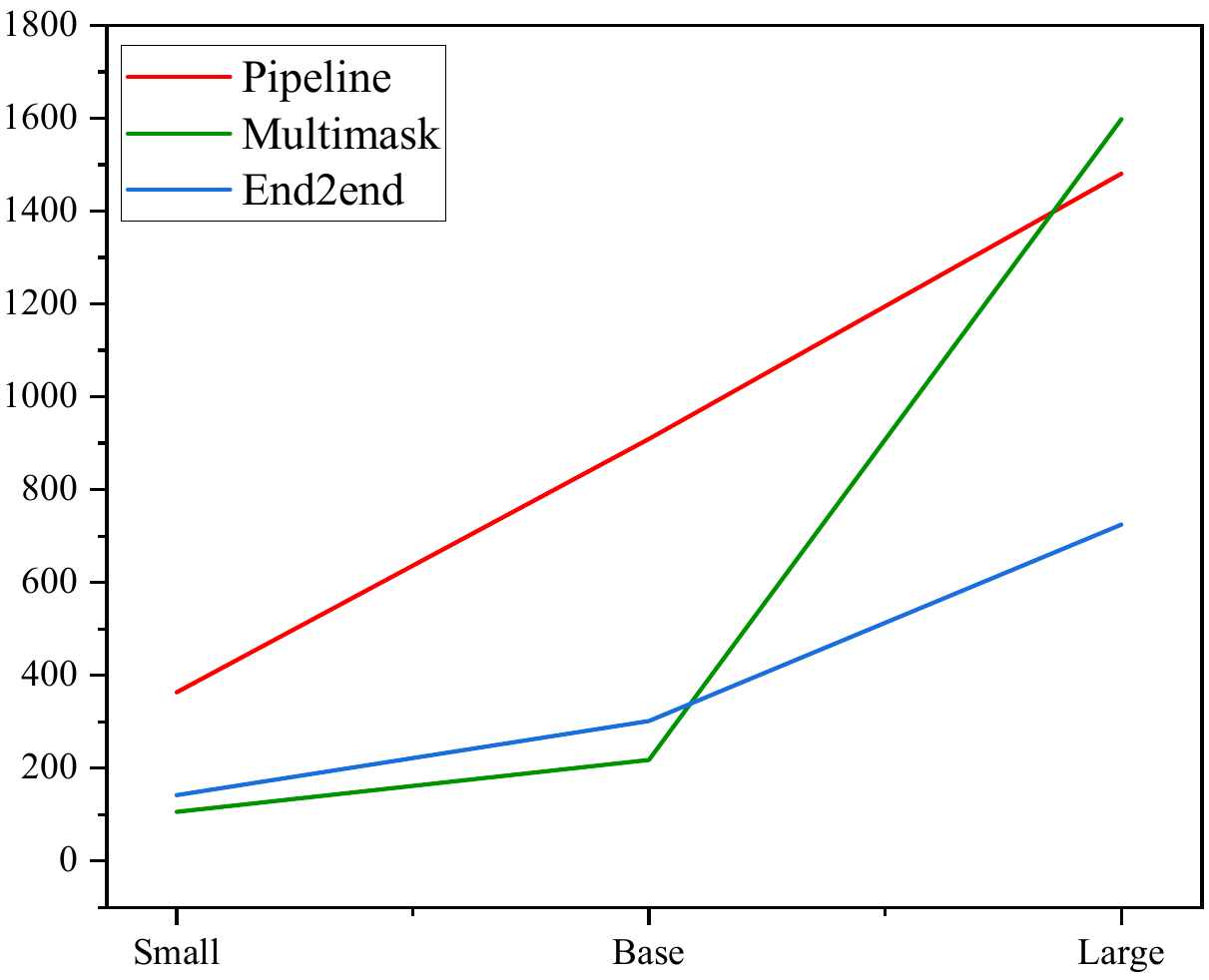}
        \caption{Inference time}
        \label{fig:compute}
    \end{minipage}
    \hfill
    \begin{minipage}[b]{0.45\textwidth}
        \centering
        \includegraphics[width=\textwidth,height=0.8\textwidth]{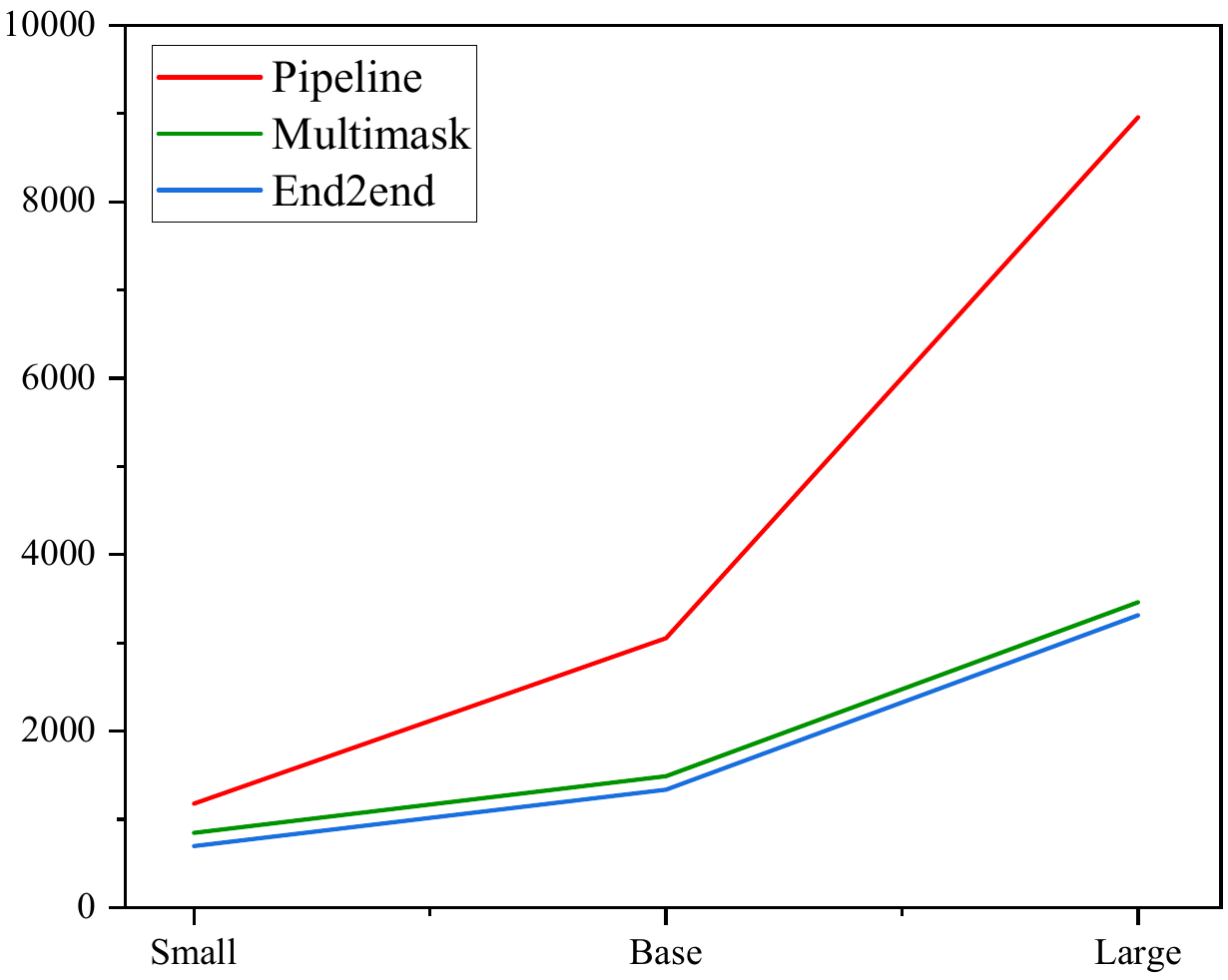}
        \caption{Memory usage}
        \label{fig:memory}
    \end{minipage}
\end{figure}

\section{Conclusion}
In this paper, we take QOG as a task for generating questions and options given an input context, and train QOG models using three different approaches. 
To evaluate them, we propose two approaches:
First, we use the QOG models to generate QO pairs on different domains and use $F_1$ scores to assess the generalisation ability of the QOG models to other domains; 
Second, GPT-4 is utilised to select the correct options in the QOs, with higher GPT-4 scores mean better question quality.
The evaluation shows that the End2end QOG model is not only the fastest to generate, but also the most effective. The findings of this study are encouraging,
as the quality of the QO datasets generated by the QOG model is close to that generated by Llama 3, and they consume much less resources than the large language model.

\section*{Limitations}
For our research, the input to our QOG models is limited to 1024 tokens, for longer texts it will not work. 
The options used for training are short and the logic for answering questions is simple, as a result, 
our models cannot be used to generate longer and more complex options as well.The models are only applicable to English scenarios, 
if we want to apply them to other languages, we need a multilingual QO dataset for training and evaluation. We will try our best to solve these limitations in the future.



\bibliography{acml24}

\clearpage

\appendix

\section{Hyper Parameters}\label{apd:first}

In fine-tuning each QOG model, we searched for the optimal hyperparameters, and \autoref{tab:apd1} shows the best hyperparameters.
The maximum input length is fixed as 512, and the maximum output length is 256.

\begin{table}[h]
    \centering
    \renewcommand{\arraystretch}{1.5}
    \resizebox{0.8\textwidth}{!}{
    \begin{tabular}{@{}l@{\hspace{3pt}}p{3cm}@{\hspace{5pt}}p{2cm}@{\hspace{5pt}}p{2cm}@{\hspace{5pt}}p{2cm}@{\hspace{5pt}}p{2cm}@{\hspace{5pt}}p{2cm}@{\hspace{5pt}}p{2cm}@{}}
    \toprule
    \multicolumn{2}{@{}l}{Approach}     & Model                   & Epoch     & LR        & Batch   \\\midrule
                        & Pipeline(AE)  & T5\textsubscript{SMALL} & 6         & 0.0001     & 256   \\
                        & Pipeline(QG)  & T5\textsubscript{SMALL} & 9         & 0.0001     & 64   \\
                        & Pipeline(DG)  & T5\textsubscript{SMALL} & 13        & 0.00005    & 64   \\
                        & Multitask     & T5\textsubscript{SMALL} & 7         & 0.0001     & 128   \\
                        & End2end       & T5\textsubscript{SMALL} & 16        & 0.0001     & 128   \\
                        & Pipeline(AE)  & T5\textsubscript{BASE}  & 8         & 0.0001     & 64   \\
                        & Pipeline(QG)  & T5\textsubscript{BASE}  & 5         & 0.0001     & 32   \\
                        & Pipeline(DG)  & T5\textsubscript{BASE}  & 13        & 0.00005    & 32   \\
                        & Multitask     & T5\textsubscript{BASE}  & 7         & 0.0001     & 64   \\
                        & End2end       & T5\textsubscript{BASE}  & 17        & 0.0001     & 64   \\
                        & Pipeline(AE)  & T5\textsubscript{LARGE} & 9         & 0.0001     & 32   \\
                        & Pipeline(QG)  & T5\textsubscript{LARGE} & 6         & 0.00005    & 32   \\
                        & Pipeline(DG)  & T5\textsubscript{LARGE} & 8         & 0.00001    & 32   \\
                        & Multitask     & T5\textsubscript{LARGE} & 5         & 0.0001     & 32   \\
                        & End2end       & T5\textsubscript{LARGE} & 12        & 0.0001     & 32   \\\bottomrule
    \end{tabular}
  }
    \caption{Optimal hyperparameters for each QOG model.}
    \label{tab:apd1}
\end{table}

\section{Additional Results of LLM QO evaluation}\label{apd:second}

\autoref{tab:apd2} shows the QO evaluation results of some large language models with small parameters.
Qwen2-7B has the highest score among all models.
Phi-3, as a smaller parameter model, also performs well in the QO evaluation.

\begin{table}[t]
    \centering
    \renewcommand{\arraystretch}{1.5}
    \resizebox{0.8\textwidth}{!}{
    \begin{tabular}{@{}l@{\hspace{3pt}}l@{\hspace{5pt}}c@{\hspace{5pt}}c@{\hspace{5pt}}c@{\hspace{5pt}}c@{\hspace{5pt}}c@{\hspace{5pt}}c@{}}
    \toprule
    \multicolumn{2}{@{}l}{Model}  & Aver         & Amazon     & Wiki        & NYT         & Reddit     & Fin \\\midrule
                        & Qwen2-7B  & 57.98 & 47.37 & 58.72 & 55.29 & 52.51 & 76.03 \\
                        & Qwen1.5-MoE-A2.7B & 48.73 & 34.16  & 55.29 & 43.56 & 35.91  & 74.72\\
                        & Deepseek-7B   & 54.32 & 51.58 & 37.78 & 61.20 & 62.09  & 58.95\\
                        & GLM-4-9B  & 42.00 & 51.56 & 43.06 & 8.61 & 36.24  & 70.55\\
                        & Baichuan2-7B & 43.64 & 48.82 & 33.07 & 39.01 & 18.35  & 78.97\\
                        & Mistral-7B   & 38.68 & 42.86 & 34.23 & 37.32 & 44.45  & 34.54\\
                        & Gemma-7B  & 43.38 & 48.59 & 46.67 & 57.45 & 10.93 & 53.26 \\
                        & Phi-3-3.8B & 52.98 & 48.25 & 52.01 & 70.48 & 51.07 & 43.09 \\\bottomrule
    \end{tabular}
  }
    \caption{QO evaluation results($F_1$) of LLM on the test set.}
    \label{tab:apd2}
\end{table}

\section{Additional Results of GPT-4 evaluation}\label{apd:third}
We also use the GPT evaluation method to evaluate the QO generation quality of each llm.
The result is shown in \autoref{tab:apd3}.

\begin{table}[h]
    \centering
    \resizebox{0.4\textwidth}{!}{
    \begin{tabular}{@{}lc@{}}
    \toprule
    \textbf{Model}                     & \textbf{GPT-4} \\\midrule
    Qwen2-7B  & 98  \\
    Qwen1.5-MoE-A2.7B & 88 \\
    Deepseek-7B   & 95 \\
    GLM-4-9B  & 81 \\
    Baichuan2-7B & 85\\
    Mistral-7B   & 83 \\
    Gemma-7B  & 93 \\
    Phi-3-3.8B & 87 \\\bottomrule
    \end{tabular}
}
    \caption{Mean scores for each LLM under the GPT-4 judgement}
    \label{tab:apd3}
\end{table}

\end{document}